\documentclass{article} 
\usepackage{nips12submit_e,times}
\usepackage{subfigure} 
\usepackage{graphicx}
\usepackage{amsmath}
\usepackage{color}
\usepackage{amssymb}
\usepackage{algorithm}
\usepackage{algorithmic}
\usepackage{multirow}

\newcommand{\vect}[1]{\boldsymbol{#1}}

\usepackage[turnoff]{notes}

\newcommand{\PLML}{{PLML}}
\newcommand{\LMNNMM}{{LMNN-MM}}

\title{Parametric Local Metric Learning for Nearest Neighbor Classification}

\author{
Jun Wang \\
Department of Computer Science\\ 
University of Geneva\\ Switzerland\\
\texttt{Jun.Wang@unige.ch} 
\And Adam Woznica \\
Department of Computer Science\\ 
University of Geneva\\ Switzerland\\
\texttt{Adam.Woznica@unige.ch}
\And Alexandros Kalousis \\
 Department of Business Informatics\\
 University of Applied Sciences \\Western Switzerland\\
\texttt{Alexandros.Kalousis@hesge.ch} }

\nipsfinalcopy 

\begin{document}

\maketitle

\begin{abstract}

We study the problem of learning local metrics for nearest neighbor classification.
Most previous works on local metric learning learn a number of local unrelated metrics.
While this "independence" approach delivers an increased flexibility its downside is the considerable
risk of overfitting. We present a new parametric local metric learning method in which we learn a 
smooth metric matrix function over the data manifold. Using an approximation error bound of the metric 
matrix function we learn local metrics as linear combinations of basis metrics defined on anchor 
points over different regions of the instance space. 
We constrain the metric matrix
function by imposing on the linear combinations manifold regularization which makes the learned metric matrix
function vary smoothly along the geodesics of the data manifold.
Our metric learning method has excellent performance both in terms of predictive power and scalability. We 
experimented with several large-scale classification problems, tens of thousands of instances, and compared
it with several state of the art metric learning methods, both global and local, as well as to SVM with automatic
kernel selection, all of which it outperforms in a significant manner.
%
\end{abstract}

\section{Introduction}
\label{sec:intro}
The nearest neighbor (NN) classifier is one of the simplest and most classical non-linear
classification algorithms. It is guaranteed to yield an error no 
worse than twice the Bayes error as the number of instances approaches infinity. With 
finite learning instances, its performance strongly depends on the use of an appropriate distance measure. 
Mahalanobis metric learning~\cite{davis2007information,weinberger2009distance,jain2009metric,jin2009regularized,ying2009sparse,wangmetric} 
improves the performance of the NN classifier if used instead of the Euclidean metric. It learns a global distance metric which determines 
the importance of the different input features and their correlations. However, since the discriminatory power of the input features might 
vary between different neighborhoods, learning a global metric cannot fit well the distance over the data manifold. Thus a more appropriate 
way is to learn a metric on each neighborhood 
and {\em local metric learning}~\cite{hastie1996discriminant,bilenko2004integrating,weinberger2009distance,frome2007image} does exactly that. It 
increases the expressive power of standard Mahalanobis metric learning by learning a number of local metrics (e.g. one per each instance).

Local metric learning has been shown to be effective for different learning scenarios. One of the first local metric learning works, Discriminant Adaptive 
Nearest Neighbor classification~\cite{hastie1996discriminant}, DANN, learns local metrics by shrinking neighborhoods in directions orthogonal to the local 
decision boundaries and enlarging the neighborhoods parallel to the boundaries. It learns the local metrics independently with no regularization between 
them which makes it prone to overfitting.
The authors of LMNN-Multiple Metric (\LMNNMM)~\cite{weinberger2009distance} significantly limited the number of learned metrics and 
constrained all instances in a given region to share the same metric in an effort to combat overfitting. In the supervised setting 
they fixed the number of metrics to the number of classes; a similar idea has been also considered in~\cite{bilenko2004integrating}. However,
they too learn the metrics independently for each region making them also prone to overfitting since the local metrics will be overly specific to 
their respective regions. 
The authors of \cite{yeung2007locally} learn local metrics using a least-squares approach by minimizing a 
weighted sum of the distances of each instance to apriori defined target positions and constraining the instances in the
projected space to preserve the original geometric structure of the data in an effort to alleviate overfitting. 
However, the method learns the local metrics using a learning-order-sensitive propagation strategy, and depends 
heavily on the appropriate definition of the target positions for each instance, a 
task far from obvious.
\note[Removed]{There the local metrics
are constrained to be smooth, i.e. metrics in neighboring regions are forced to be similar to each other.
More precisely, the developed method called Locally Smooth Metric Learning (LSML) learns local metrics
one by one by minimizing the weighted sum of distances of each instance to its a-priori defined target
positions in the projected space induced by the local metrics; the weights were set a-priori. The 
smoothness of the learned local metric is implicitly imposed by a manifold regularization of the 
instances in the projected space. However, this method has several drawbacks. 
\begin{itemize}
\item First, it learns for each instance one local metric thus similar to DANN and GLML it does not
scale well with learning problems with large number of instances. 
\item Second, its objective function 
only considers similarity constraints; nevertheless, the discriminative ability of similarity 
constraints is broadly considered as inferior to that of large margin constraints that combine 
both similarity and dissimilarity infor mation 
\item Third, in order to derive a closed-form solu- tion for each local metric this method uses a propagation 
strategy that iteratively learns the local metrics one by one; as a result, its final solution strongly 
depends on the learning order of local metrics. Finally, this method is diffiult to initialize as it 
requires the specification of the target positions for each training instance.
\end{itemize}}
In another effort to overcome the overfitting problem of the discriminative methods~\cite{hastie1996discriminant,weinberger2009distance}, 
Generative Local Metric Learning, GLML,~\cite{nohgenerative}, propose to learn local metrics by minimizing the NN expected classification error 
under strong model assumptions.
They use the Gaussian distribution to model the learning instances of each class. However, the strong model 
assumptions might easily be very inflexible for many learning problems.

In this paper we propose the Parametric Local Metric Learning method (\PLML) which learns a {\em smooth metric matrix function}
over the data manifold.  More precisely, we parametrize the metric matrix of each instance as a linear combination of basis metric matrices 
of a small set of anchor points; this parametrization is naturally derived from an error bound 
on local metric approximation. Additionally we incorporate a manifold regularization on the linear combinations, 
forcing the linear combinations to vary smoothly over the data manifold. We develop an efficient two 
stage algorithm that first learns the linear combinations of each instance and then the metric matrices 
of the anchor points.
To improve scalability and efficiency we employ a fast first-order optimization algorithm, FISTA~\cite{beck2010gradient}, 
to learn the linear combinations as well as the basis metrics of the anchor points.
We experiment with the \PLML\ method on a number of large scale classification problems with tens of 
thousands of learning instances. 
The experimental results clearly demonstrate that \PLML\ significantly improves the 
predictive performance over the current state-of-the-art metric learning methods, as 
well as over multi-class SVM with automatic kernel selection.


\section{Preliminaries}
\label{sec:Pre}
We denote by $\mathbf X$ the $n \times d$ matrix  of learning instances, the $i$-th row of which is the 
$\mathbf x_i^T \in \mathbb{R}^d$ instance, and by $\mathbf y=(y_1,\ldots, y_n)^T$, $y_i \in \{1,\ldots,c\}$ the
vector of class labels. The squared Mahalanobis distance between two instances in the input space is given by:
\begin{eqnarray*}
d^2_{\mathbf{M}}(\mathbf{x}_i, \mathbf{x}_j)=(\mathbf{x}_i-\mathbf{x}_j)^T\mathbf{M}(\mathbf{x}_i-\mathbf{x}_j)
\end{eqnarray*}
where $\mathbf{M}$ is a PSD metric matrix ($\mathbf{M} \succeq 0$). 
A linear metric learning method learns a Mahalanobis metric $\mathbf{M}$ by optimizing some cost function under the PSD constraints for $\mathbf{M}$ 
and a set of additional constraints on the pairwise instance distances. Depending on the actual metric learning method, different kinds of constraints 
on pairwise distances are used. The most successful ones are the large margin triplet constraints. A triplet constraint denoted by $c(\mathbf x_i,\mathbf x_j,\mathbf x_k)$, 
indicates that in the projected space induced by $\mathbf{M}$ the distance between $\mathbf x_i$ and $\mathbf x_j$ should be smaller than the distance 
between $\mathbf x_i$ and $\mathbf x_k$. 
 
Very often a single metric $\mathbf{M}$ can not model adequately the complexity of a given learning problem in which discriminative features vary between different 
neighborhoods. To address this limitation in local metric learning we learn a set of local metrics. 
In most cases we learn a local metric for each learning instance~\cite{hastie1996discriminant,nohgenerative}, 
however we can also learn a local metric for some part of the instance space in which case the number of learned metrics
can be considerably smaller than $n$, e.g.~\cite{weinberger2009distance}.  We follow the former approach and learn one 
local metric per instance. 
In principle, distances should then be defined as geodesic distances using
the local metric on a Riemannian manifold. However, this is computationally difficult, thus we define the distance between instances $\mathbf x_i$ and $\mathbf x_j$ as:
\begin{eqnarray*}
d^2_{\mathbf{M}_i}(\mathbf{x}_i, \mathbf{x}_j)=(\mathbf{x}_i-\mathbf{x}_j)^T\mathbf{M}_i(\mathbf{x}_i-\mathbf{x}_j)
\end{eqnarray*}
where $\mathbf M_i$ is the local metric of instance $\mathbf x_i$. 
Note that most often the local metric $\mathbf M_i$ of instance $\mathbf x_i$ is different from that of $\mathbf x_j$. As a result, the 
distance $d^2_{\mathbf{M_i}}(\mathbf{x}_i, \mathbf{x}_j)$ does not satisfy the symmetric property, i.e. it is not a proper metric. Nevertheless, 
in accordance to the standard practice we will continue to use the term local metric learning following \cite{weinberger2009distance,nohgenerative}.

\section{Parametric Local Metric Learning}
\label{sec:LSLMML}
We assume that there exists a Lipschitz smooth vector-valued function $f(\mathbf x)$, the output of which is the vectorized local metric matrix of 
instance $\mathbf x$. Learning the local metric of each instance is essentially learning the value of this function at different points over the 
data manifold. In order to significantly reduce the computational complexity we will approximate the metric function instead of directly learning it.

\newtheorem{definition}{Definition}
\begin{definition}
\label{Lipschitz}
A vector-valued function $f(\mathbf x)$ on $\mathbb{R}^d$ is a $(\alpha, \beta, p)$-Lipschitz smooth function with respect to a 
vector norm $\left\|\cdot\right\|$ if $\left\|f(\mathbf x)-f(\mathbf x')\right\| \leq \alpha \left\|\mathbf x-\mathbf x'\right\|$ 
and  $\left\|f(\mathbf x)-f(\mathbf x')-\nabla f(\mathbf x')^T(\mathbf x-\mathbf x')\right\| \leq \beta\left\|\mathbf x-\mathbf x'\right\|^{1+p} $, 
where $\nabla f(\mathbf x')^T$ is the derivative of the $f$ function at $\mathbf x'$. We assume $\alpha, \beta > 0$ and $p \in (0,1]$.
\end{definition}

\cite{yu2009nonlinear} have shown that any Lipschitz smooth real function $f(\mathbf x)$ defined on a lower dimensional manifold can be approximated 
by a linear combination of function values $f(\mathbf u), \mathbf u \in \mathbf U$, of a set $\mathbf U$ of anchor points. Based on this result we 
have the following lemma that gives the respective error bound for learning a Lipschitz smooth vector-valued function.
\newtheorem{lemma}{Lemma}
\begin{lemma}
\label{LinAppBound}
Let $(\vect \gamma,\mathbf U)$ be a nonnegative weighting on anchor points $\mathbf U$ in $\mathbb{R}^d$. Let $f$ be an $(\alpha, \beta, p)$-Lipschitz smooth vector
function. We have for all $\mathbf x \in \mathbb{R}^d$:
\begin{eqnarray}
\label{bound}
&&\left\|f(\mathbf x)-\sum_{\mathbf u \in \mathbf U}\gamma_{\mathbf u}(\mathbf x)f(\mathbf u) \right\|\leq 
\alpha \left\|\mathbf x-\sum_{\mathbf u \in \mathbf U}\gamma_{\mathbf u}(\mathbf x)\mathbf u  \right\|+\beta \sum_{\mathbf u \in \mathbf U} \gamma_{\mathbf u}(\mathbf x) \left\|\mathbf x- \mathbf u  \right\|^{1+p}
\end{eqnarray}
\end{lemma}
The proof of the above Lemma \ref{LinAppBound} is similar to the proof of Lemma 2.1 in~\cite{yu2009nonlinear}; for lack of space we omit its presentation. By the nonnegative weighting strategy $(\vect \gamma,\mathbf U)$, the PSD constraints on the approximated local metric is automatically satisfied if the local metrics of anchor points are PSD matrices. 

Lemma \ref{LinAppBound} suggests a natural way to approximate the local metric function by parameterizing the metric $\mathbf M_i$ of each 
instance $\mathbf x_i$ as a {\em weighted linear combination}, $\mathbf W_i \in \mathbb{R}^m$, of a small set of metric basis, $\{\mathbf M_{b_1},\ldots,\mathbf{M}_{b_m}\}$, 
each one associated with an anchor point defined in some region of the instance space. This parametrization will also  provide us with a global way to 
regularize the flexibility of the metric function. We will first learn the vector of weights $\mathbf{W}_i$ for each 
instance $\mathbf x_i$, and then the basis metric matrices; these two together, will give us the $\mathbf M_i$ metric for the instance $\mathbf x_i$. 
\note[Alexandros]{I am reading all the time the reference to the~\cite{yu2009nonlinear} it seems as if all the work is taken from there.
Would it make sense to remove some of them?}

More formally, we define a $m \times d$ matrix $\mathbf U$ of anchor points, the $i$-th row of which is the anchor point $\mathbf u_i$, where $\mathbf u_i^T \in \mathbb{R}^d$. 
We denote by $\mathbf M_{b_i}$ the Mahalanobis metric matrix associated with $\mathbf u_i$. The anchor points can be defined using some clustering algorithm, 
we have chosen to define them as the means of clusters constructed by the $k$-means algorithm. The local metric $\mathbf M_i$ of an instance $\mathbf x_i$ is 
parametrized by:
\begin{eqnarray}
\label{localM-parametrization}
\mathbf M_i=\sum_{b_k} W_{i{b_k}} \mathbf M_{b_k},\ W_{i{b_k}} \geq 0,\ \sum_{b_{k}} W_{i{b_k}}=1 
\end{eqnarray}
where $\mathbf W$ is a $n \times m$ weight matrix, and its $W_{ib_k}$ entry is the weight of the basis metric  $\mathbf M_{b_k}$ for the instance $\mathbf x_i$. The constraint $\sum_{b_k}W_{ib_k}=1$ removes
 the scaling problem between different local metrics.
%
Using the parametrization of equation (\ref{localM-parametrization}), the squared distance of $\mathbf x_i$ to $\mathbf x_j$ under the metric $\mathbf M_i$ is:
\begin{eqnarray}
\label{pairwise.general}
&&d_{\mathbf M_i}^2(\mathbf{x}_i,\mathbf{x}_j)=\sum_{b_k} W_{i{b_k}} d^2_{\mathbf M_{b_k}}(\mathbf{x}_i,\mathbf{x}_j) \\\nonumber
\end{eqnarray}
where $d^2_{\mathbf M_{b_k}}(\mathbf{x}_i,\mathbf{x}_j)$ is the squared Mahalanobis distance between $\mathbf x_i$ and $\mathbf x_j$ under 
the basis metric $\mathbf M_{b_k}$. We will show in the next section 
how to learn the weights of the basis metrics for each instance  and in section~\ref{sec:LLMML} how to learn the basis metrics. 

\subsection{Smooth Local Linear Weighting}
\label{sec:SNLLW}
Lemma~\ref{LinAppBound} bounds the approximation error by two terms. 
The first term states that $\mathbf x$ should be close to its linear approximation, 
and the second that the weighting should be local. In addition we want the local metrics 
to vary smoothly over the data manifold. To achieve this smoothness we rely on manifold 
regularization and constrain the weight vectors of neighboring instances to be similar.
Following this reasoning we will learn Smooth Local Linear 
Weights for the basis metrics by minimizing the error bound 
of (\ref{bound}) together with a regularization term that controls 
the weight variation of similar instances. To simplify the objective function, 
\note[Alexandros]{Again the reference to ~\cite{yu2009nonlinear}, can we not remove it?}
we use the term $\left\|\mathbf x-\sum_{\mathbf u \in 
\mathbf U}\gamma_{\mathbf u}(\mathbf x)\mathbf u  \right\|^2$ instead of $\left\|\mathbf x-\sum_{\mathbf u \in \mathbf U}\gamma_{\mathbf u}(\mathbf x)\mathbf u  \right\|$.
By including the constraints on the $\mathbf W$ weight matrix in (\ref{localM-parametrization}), the optimization problem is given by:
\begin{eqnarray}
\label{weight-learning-opt}
\min_{\mathbf W}g(\mathbf W)&=&\left\|\mathbf X-\mathbf {WU}\right\|^2_{F}+\lambda_1 tr(\mathbf {WG})
+\lambda_2 tr(\mathbf {W^TLW})\\\nonumber
s.t.&& \mathbf W_{i{b_k}} \geq 0,\sum_{b_k}\mathbf W_{i{b_k}}=1, \forall i, b_k\\\nonumber
\end{eqnarray}
where $tr(\cdot)$ and $\left\|\cdot \right\|_{F}$ denote respectively the trace norm of a square matrix and the 
Frobenius norm of a matrix. The $m \times n$ matrix $\mathbf G$ is the squared distance matrix between each 
anchor point $\mathbf u_{i}$ and each instance $\mathbf x_j$, obtained for $p=1$ in (\ref{bound}), i.e. its $(i,j)$ entry is the squared Euclidean 
distance between $\mathbf u_{i}$ and $\mathbf x_j$. $\mathbf L$ is the $n \times n$ Laplacian matrix constructed by $\mathbf D-\mathbf S$, where $\mathbf S$ 
is the $n \times n$ symmetric pairwise similarity matrix of learning instances and $\mathbf D$ is a diagonal matrix with $\mathbf D_{ii}=\sum_{k}\mathbf S_{ik}$. 
Thus the minimization of the  $tr(\mathbf {W^TLW})$ term constrains similar instances to have similar weight coefficients. The minimization of 
the $tr(\mathbf {WG})$ term forces the weights of the instances to reflect their local properties. Most often the similarity matrix $\mathbf S$ 
is constructed using $k$-nearest neighbors graph \cite{zelnik2004self}. The $\lambda_1$ and $\lambda_2$ parameters control the importance of the different terms.

Since the cost function $g(\mathbf W)$ is convex quadratic with $\mathbf W$ and the constraint is simply linear, (\ref{weight-learning-opt}) is a convex 
optimization problem with a unique optimal solution.  The constraints on $\mathbf W$ in (\ref{weight-learning-opt}) can be seen as $n$ simplex constraints 
on each row of $\mathbf W$; we will use the projected gradient method to solve the optimization problem. At each iteration $t$, the learned weight 
matrix $\mathbf W$ is updated by: 
\begin{eqnarray}
\label{weight-upt}
\mathbf W^{t+1}=Proj(\mathbf W^{t}-\eta \nabla g(\mathbf W^{t}))
\end{eqnarray}
where $\eta > 0$ is the step size and $\nabla g(\mathbf W^{t})$ is the gradient of the 
cost function $g(\mathbf W)$ at $\mathbf W^t$. The $Proj(\cdot)$ denotes the simplex projection operator
on each row of $\mathbf W$. Such a projection operator can be efficiently implemented with a complexity of $O(nm\log(m))$~\cite{duchi2008efficient}. 
\begin{algorithm}[t]
   \caption{Smoothl Local Linear Weight Learning}
   \label{algo:SLLW}
   \begin{algorithmic}
   \STATE {\bfseries Input:} $\mathbf{W^0}$, $\mathbf{X}$, $\mathbf{U}$, $\mathbf{G}$, $\mathbf{L}$, $\lambda_1$, and $\lambda_2$
   \STATE {\bfseries Output:} matrix $\mathbf{W}$
   \STATE define  $\widetilde{g}_{\beta,\mathbf Y}(\mathbf W)=g(\mathbf Y)+tr(\nabla g(\mathbf Y)^T(\mathbf {W-Y}))+\frac{\beta}{2} \left\|\mathbf{W-Y}\right\|^2_F$
   \STATE initialize: $t_1=1$, $\beta=1$,$\mathbf Y^1=\mathbf W^0$, and $i=0$
   \REPEAT
   \STATE $i=i+1$, $\mathbf W^i=Proj((\mathbf Y^{i}-\frac{1}{\beta} \nabla g(\mathbf Y^{i})))$
   \WHILE {$g(\mathbf W^i) > \widetilde{g}_{\beta,\mathbf Y^i}(\mathbf W^i)$}
   \STATE $\beta=2\beta$, $\mathbf W^i=Proj((\mathbf Y^{i}-\frac{1}{\beta} \nabla g(\mathbf Y^{i})))$ 
   \ENDWHILE
    \STATE $t_{i+1}=\frac{1+\sqrt{1+4t_{i}^2}}{2}$, $\mathbf Y^{i+1}=\mathbf W^i +\frac{t_i-1}{t_{i+1}}(\mathbf W^i-\mathbf W^{i-1})$
   \UNTIL {converges;}
\end{algorithmic}
\end{algorithm}
To speed up the optimization procedure we employ a fast first-order optimization method FISTA,~\cite{beck2010gradient}. 
The detailed algorithm is described 
in Algorithm~\ref{algo:SLLW}. The Lipschitz constant $\beta$ required by this algorithm is estimated by using the condition of $g(\mathbf W^i) \leq \widetilde{g}_{\beta,\mathbf Y^i}(\mathbf W^i)$~\cite{bachconvex}. At each iteration, the main computations are in the gradient and the objective value with complexity $O(nmd+n^2m)$.

To set the weights of the basis metrics for a testing instance we can optimize (\ref{weight-learning-opt}) given the weight of the basis metrics for the training instances. 
Alternatively we can simply set them as the weights of its nearest neighbor in the training instances. In the experiments we used the latter approach.

\subsection{Large Margin Basis Metric Learning}
\label{sec:LLMML}
In this section we define a large margin based algorithm to learn the basis metrics $\mathbf M_{b_1},\ldots,\mathbf{M}_{b_m}$.
Given the $\mathbf W$ weight matrix of basis metrics obtained using Algorithm~\ref{algo:SLLW}, the local metric
$\mathbf M_{i}$ of an instance $\mathbf x_i$ defined in (\ref{localM-parametrization}) is linear with respect to the basis 
metrics $\mathbf M_{b_1},\ldots,\mathbf{M}_{b_m}$. We define the relative comparison distance of instances $\mathbf x_i$, 
$\mathbf x_j$ and $\mathbf x_k$ as:
$d_{\mathbf M_i}^2(\vect{x}_i,\vect{x}_k)-d_{\mathbf M_i}^2(\vect{x}_i,\vect{x}_j)$.
In a large margin constraint $c(\mathbf x_i,\mathbf x_j,\mathbf x_k)$, the squared distance $d_{\mathbf M_i}^2(\vect{x}_i,\vect{x}_k)$ is required to be 
larger than $d_{\mathbf M_i}^2(\vect{x}_i,\vect{x}_j)+1$, otherwise an error $\vect \xi_{ijk} \geq 0$ is generated. 
Note that, this relative comparison definition 
is different from that defined in LMNN-MM \cite{weinberger2009distance}. In LMNN-MM to avoid over-fitting, different local metrics $\mathbf M_j$ and $\mathbf M_k$ 
are used to compute the squared distance $d_{\mathbf M_j}^2(\vect{x}_i,\vect{x}_j)$ and $d_{\mathbf M_k}^2(\vect{x}_i,\vect{x}_k)$ respectively, as no smoothness constraint is added between metrics of different local regions. 

Given a set of triplet constraints, we learn the basis metrics $\mathbf M_{b_1},\ldots,\mathbf{M}_{b_m}$ with the following optimization problem:
\begin{eqnarray}
\label{opt.MLM}
\min_{\mathbf M_{b_1},\ldots,\mathbf{M}_{b_m},\vect \xi}&& \alpha_1 \sum_{b_l} ||\mathbf M_{b_l}||^2_F +\sum_{ijk} \vect \xi_{ijk}
 +\alpha_2 \sum_{ij}\sum_{b_l} W_{ib_l}d_{\mathbf M_{b_l}}^2(\vect{x}_i,\vect{x}_j)\\\nonumber
s.t. &&\sum_{b_l} W_{ib_l}(d_{\mathbf M_{b_l}}^2(\vect{x}_i,\vect{x}_k)-d_{\mathbf M_{b_l}}^2(\vect{x}_i,\vect{x}_j)) 
\geq 1 - \vect \xi_{ijk} \ \forall {i, j, k} \\\nonumber
&&\vect \xi_{ijk} \geq 0; \ \forall {i, j, k} \
\mathbf M_{b_l} \succeq 0;\ \forall {b_l} \nonumber
\end{eqnarray}
where $\alpha_1$ and $\alpha_2$ are parameters that balance the importance of the different terms. The large margin triplet constraints for each instance are 
generated using its $k_1$ same class nearest neighbors and $k_2$ different class nearest neighbors by requiring its distances to the $k_2$ different class 
instances to be larger than those to its $k_1$ same class instances.  In the objective function of (\ref{opt.MLM}) the basis metrics are learned by minimizing 
the sum of large margin errors and the sum of squared pairwise distances of each instance to its $k_1$ nearest neighbors computed using the local metric. 
Unlike LMNN we add the squared Frobenius norm on each basis metrics in the objective function. We do this for two reasons. First we exploit the connection 
between LMNN and SVM shown in~\cite{do2012metric} under which the squared Frobenius norm of the metric matrix is related to the SVM margin. Second because adding 
this term leads to an easy-to-optimize dual formulation of (\ref{opt.MLM})~\cite{shen2011scalable}.


Unlike many special solvers which optimize the primal form of the metric learning problem~\cite{weinberger2009distance,shen2011positive}, we follow~\cite{shen2011scalable} 
and optimize the Lagrangian dual problem of (\ref{opt.MLM}). The dual formulation leads to an efficient basis metric learning algorithm. Introducing the Lagrangian dual 
multipliers $\vect \gamma_{ijk}$, $\vect p_{ijk}$ and the PSD matrices $\mathbf Z_{b_l}$ 
to respectively associate with every large margin triplet constraints, $\vect \xi_{ijk} \geq 0$ 
and the PSD constraints $\mathbf M_{b_l} \succeq 0$ in (\ref{opt.MLM}), we can easily derive the following Lagrangian dual form
\begin{eqnarray}  
\label{dualform}
\max_{\mathbf Z_{b_1}, \ldots, \mathbf Z_{b_m},\vect \gamma} && \sum_{ijk}\gamma_{ijk}-\sum_{b_l}\frac{1}{4\alpha_1}\cdot\|\mathbf Z_{b_l}+
\sum_{ijk}\gamma_{ijk} W_{ib_l} \mathbf C_{ijk}-\alpha_2 \sum_{ij} W_{ib_l} \mathbf A_{ij}\|_F^2\\\nonumber
s.t. && 1 \geq \gamma_{ijk} \geq 0; \ \forall_{i, j, k} \
 \mathbf Z_{b_l} \succeq 0;\ \forall {b_l} \nonumber
\end{eqnarray}
and the corresponding optimality conditions:
$\mathbf M_{b_l}^*=\frac{(\mathbf Z_{b_l}^*+ \sum_{ijk}\gamma_{ijk}^* W_{i{b_l}} \mathbf C_{ijk}-\alpha_2 \sum_{ij} W_{i{b_l}} \mathbf A_{ij})}{2\alpha_1}$
and
$1 \geq \gamma_{ijk} \geq 0$,
where the matrices $\mathbf A_{ij}$ and $\mathbf C_{ijk}$ are  given by
 $\mathbf x_{ij}^T \mathbf x_{ij}$ and $\mathbf x_{ik}^T\mathbf x_{ik}-\mathbf x_{ij}^T\mathbf x_{ij}$ respectively, where 
$\mathbf x_{ij}=\mathbf x_i-\mathbf x_j$. 

Compared to the primal form, the main advantage of the dual formulation is that the second term in the objective function of 
(\ref{dualform}) has a closed-form solution for $\mathbf Z_{b_l}$ given a fixed $\vect \gamma$. To drive the optimal solution 
of $\mathbf Z_{b_l}$, let $\mathbf K_{b_l}=\alpha_2 \sum_{ij} W_{ib_l} \mathbf A_{ij}-\sum_{ijk}\gamma_{ijk} W_{ib_l} \mathbf C_{ijk}$. 
Then, given a fixed $\vect \gamma$, the optimal solution of $\mathbf Z_{b_l}$ is
$\mathbf Z_{b_l}^*=(\mathbf K_{b_l})_+$, where $(\mathbf K_{b_l})_+$ projects the matrix $\mathbf K_{b_l}$ onto the PSD cone, i.e. $(\mathbf K_{b_l})_+= \mathbf{U[\max(diag(\Sigma)),0)]U^T}$ with $\mathbf K_{b_l}=\mathbf {U \Sigma U^T}$. 

Now, (\ref{dualform}) is rewritten as:
\begin{eqnarray}  
\label{simpledual}
\min_{\vect \gamma} && g(\vect \gamma)=-\sum_{ijk}\gamma_{ijk}
 +\sum_{b_l}\frac{1}{4\alpha_1}\left\|(\mathbf K_{b_l})_+-\mathbf K_{b_l} \right\|_F^2\\\nonumber
s.t. && 1 \geq \gamma_{ijk} \geq 0; \forall {i, j, k} \nonumber
\end{eqnarray}
And the optimal condition for $\mathbf M_{b_l}$ is $\mathbf M_{b_l}^*=\frac{1}{2\alpha_1}((\mathbf K_{b_l}^*)_+-\mathbf K_{b_l}^* )$.
The gradient of the objective function in (\ref{simpledual}), $\nabla g(\gamma_{ijk})$, is given by:
$\nabla g(\gamma_{ijk})=-1+\sum_{b_l}\frac{1}{2\alpha_1}\left\langle(\mathbf K_{b_l})_+-\mathbf K_{b_l},  W_{ib_l}\mathbf C_{ijk} \right\rangle$.
At each iteration, $\vect \gamma$ is updated by: 
\begin{eqnarray}
\label{gamma-upt}
\mathbf \gamma^{i+1}=BoxProj(\vect \gamma^{i}-\eta \nabla g(\vect \gamma^{i}))
\end{eqnarray}
where $\eta > 0$ is the step size. The $BoxProj(\cdot)$ denotes the simple box projection operator on $\vect \gamma$ as specified in the constraints of (\ref{simpledual}).
At each iteration, the main computational complexity lies in the computation of the eigen-decomposition 
with a complexity of $O(md^3)$ and the computation of the gradient with a complexity of $O(m(nd^2+cd))$, where $m$ is the number of basis metrics and $c$ is 
the number of large margin triplet constraints. As in the weight learning problem the FISTA algorithm is employed to accelerate the optimization process; 
for lack of space we omit the algorithm presentation.
 

\section{Experiments}
\label{sec:exp}

In this section we will evaluate the performance of \PLML\ and compare it with a number 
of relevant baseline methods on six datasets with large number of instances, ranging from 
5K to 70K instances; these datasets are Letter, USPS, Pendigits, Optdigits, Isolet and MNIST.
We want to determine whether the addition of manifold regularization on the local metrics improves the predictive performance of 
local metric learning, and whether the local metric learning improves over learning with single global metric. 
We will compare \PLML\ against six baseline methods. The first, SML, is a variant of \PLML\ 
where a single global metric is learned, i.e. we set the number of basis in 
(\ref{opt.MLM}) to one. The second, Cluster-Based LML (CBLML), is also a variant of \PLML\ without weight learning.
Here we learn one local metric for each cluster and we assign a weight of one for a basis metric $\mathbf M_{b_i}$ 
if the corresponding cluster of $\mathbf M_{b_i}$ contains the instance, and zero otherwise. Finally, we also compare 
against four state of the art metric learning methods LMNN~\cite{weinberger2009distance}, 
BoostMetric~\cite{shen2011positive}\footnote{http://code.google.com/p/boosting}, GLML~\cite{nohgenerative} 
and LMNN-MM~\cite{weinberger2009distance}\footnote{http://www.cse.wustl.edu/$\sim$kilian/code/code.html.}. The former two learn 
a single global metric and the latter two a number of local metrics. In addition to the different metric learning methods, we 
also compare \PLML\ against multi-class SVMs in which we use the 
one-against-all strategy to determine the class label for multi-class problems and select the best kernel with inner cross validation.

Since metric learning is computationally expensive for datasets with large number of features we followed~\cite{weinberger2009distance} 
and reduced the dimensionality of the  USPS, Isolet and MINIST datasets by applying PCA. In these datasets the retained PCA components 
explain 95\% of their total variances. We preprocessed all datasets by first standardizing the input features, and then normalizing the instances to so that their L2-norm is one.

\PLML\ has a number of hyper-parameters. To reduce the computational time we do not tune $\lambda_1$ 
and $\lambda_2$ of the weight learning optimization problem (\ref{weight-learning-opt}), and we set them to their
default values of $\lambda_1=1$ and $\lambda_2=100$. 
The Laplacian matrix $\mathbf L$ is constructed using the six nearest neighbors graph following~\cite{zelnik2004self}. The anchor 
points $\mathbf U$ are the means of clusters constructed with k-means clustering. The number $m$ of anchor points, i.e. the 
number of basis metrics, depends on the complexity of the learning problem. More complex problems will often require a larger number of 
anchor points to better model the complexity of the data. As the number of classes in the examined datasets is 10 or 26, we simply set 
$m=20$ for all datasets. In the basis metric learning problem (\ref{opt.MLM}), the number of the dual parameters $\vect \gamma$ is the 
same as the number of triplet constraints. To speedup the learning process, the triplet constraints are constructed only using the three 
same-class and the three different-class nearest neighbors for each learning instance. The parameter $\alpha_2$ is set 
to $1$, while the parameter $\alpha_1$ is the only parameter that we select from the set $\{0.01,0.1,1,10,100\}$ using 2-fold inner
cross-validation. The above setting of basis metric learning for \PLML\ is also used with the SML and CBLML methods. 
For LMNN and LMNN-MM we use their default settings, \cite{weinberger2009distance}, in which the triplet constraints 
are constructed by the three nearest same-class neighbors and all different-class samples. As a result, the number 
of triplet constraints optimized in LMNN and LMNN-MM is much larger than those of \PLML, SML, BoostMetric and CBLML.  
The local metrics are initialized by identity matrices. As in ~\cite{nohgenerative}, GLML uses the Gaussian distribution to model the learning instances from the
same class. Finally, we use the 1-NN rule to evaluate the performance of the different metric learning methods.
In addition as we already mentioned we also compare against multi-class SVM. Since the performance of the latter depends heavily on the
kernel with which it is coupled we do automatic kernel selection with inner cross validation to select the best kernel and parameter setting.
The kernels were chosen from the set of linear, polynomial (degree 2,3 and 4), and Gaussian kernels; the width of the Gaussian kernel was set 
to the average of all pairwise distances. Its $C$ parameter of the hinge loss term was selected from $\{0.1,1,10,100\}$.

\begin{figure}[t]
  \centering
  \subfigure[LMNN-MM]{\includegraphics[width=0.2\textwidth]{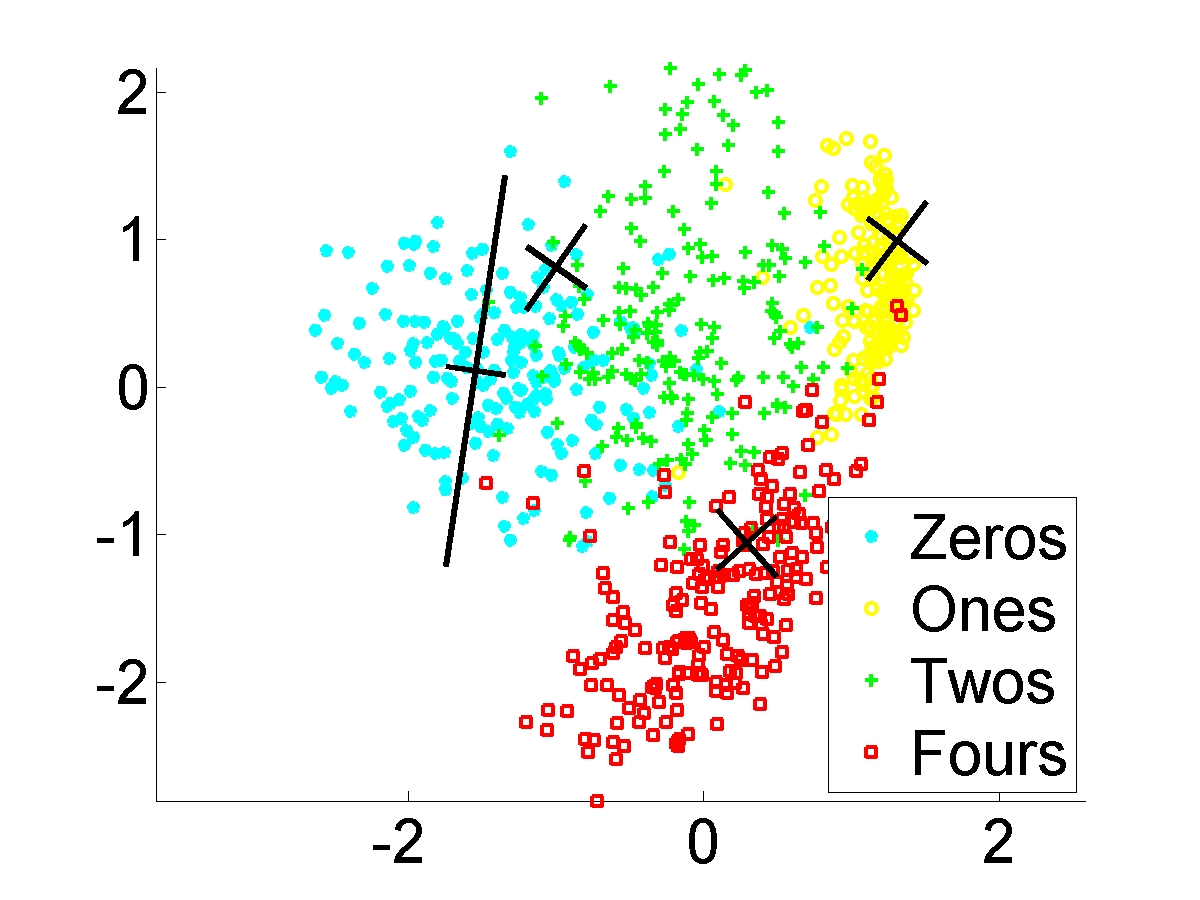}}
  \subfigure[CBLML]{\includegraphics[width=0.2\textwidth]{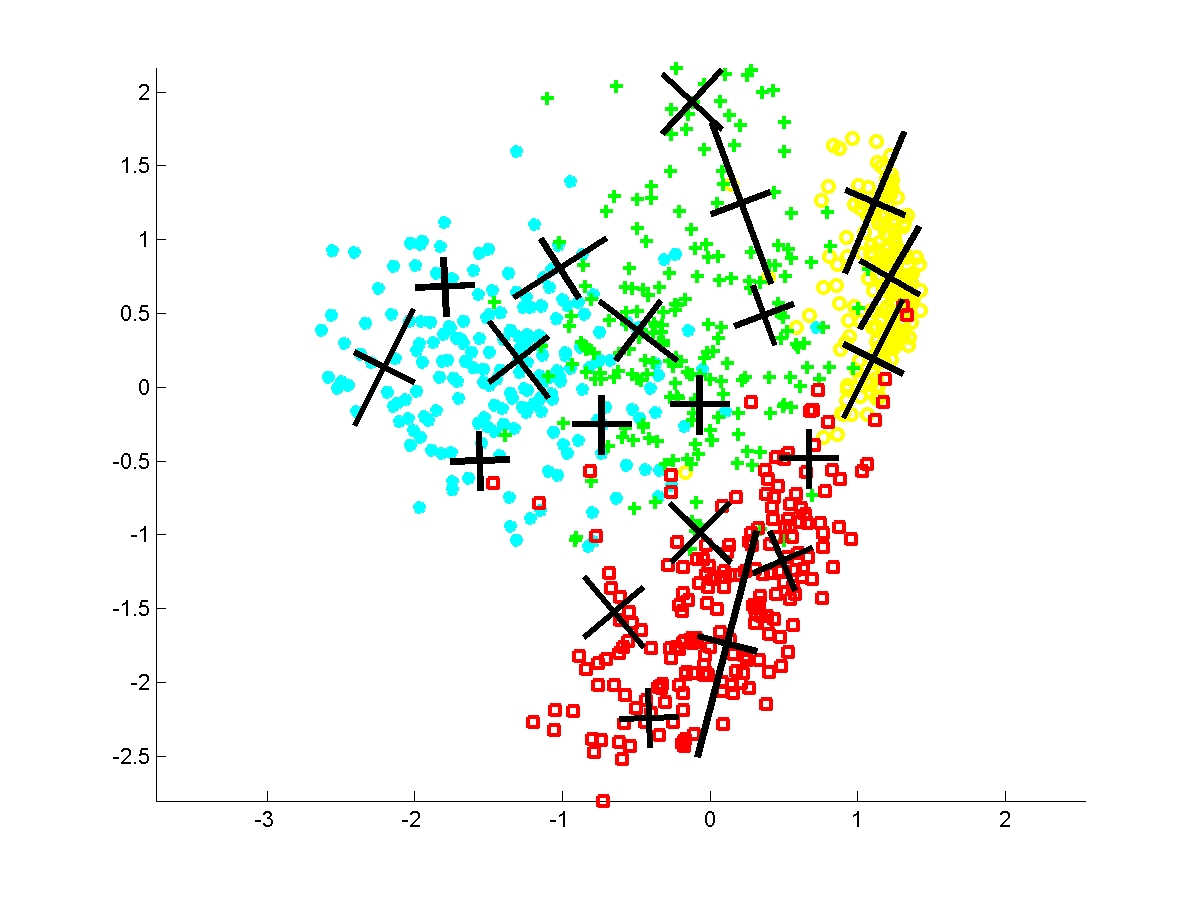}}
  \subfigure[GLML]{\includegraphics[width=0.2\textwidth]{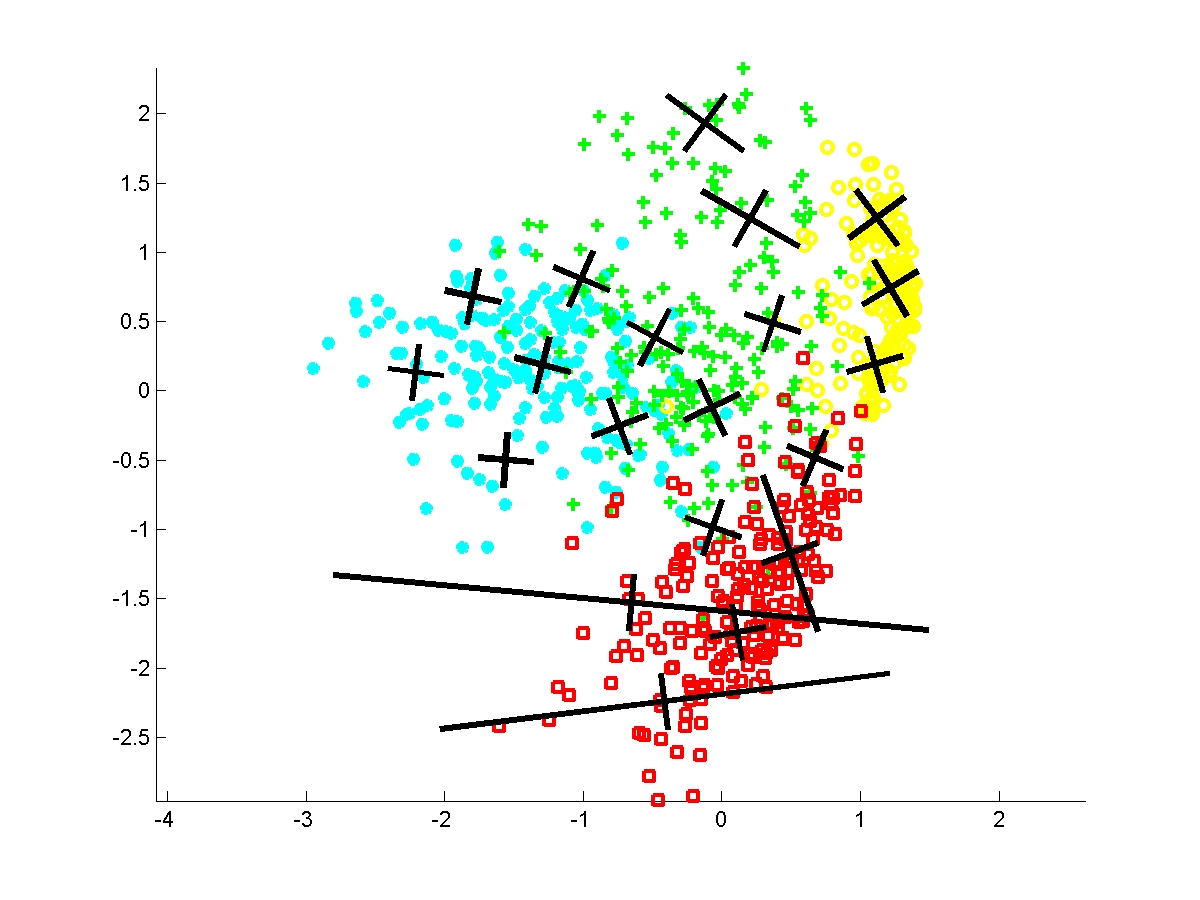}}
  \subfigure[PLML]{\includegraphics[width=0.2\textwidth]{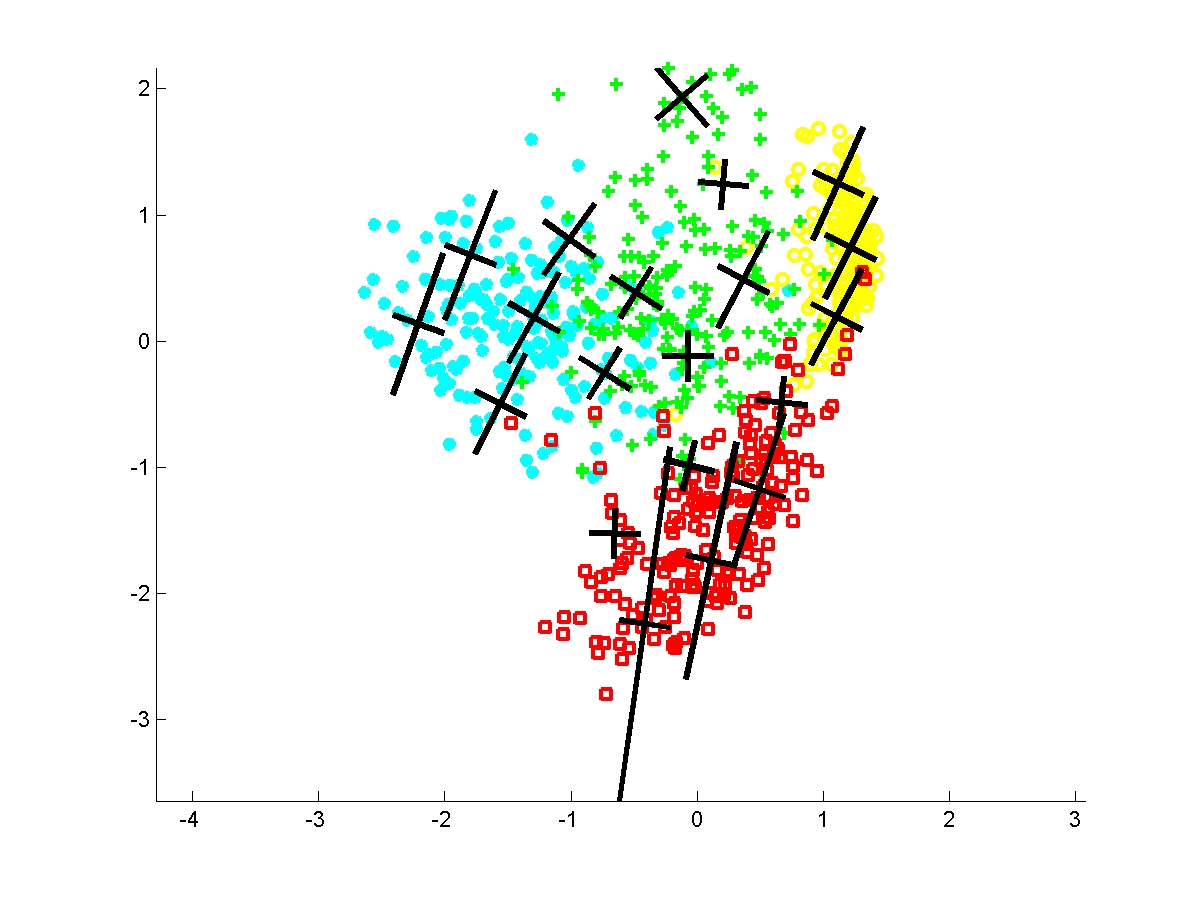}}
  \caption{The visualization of learned local metrics of LMNN-MM, CBLML, GLML and PLML. }
\label{visulization}
\end{figure}

\begin{table}[bt!]
\begin{center}
\caption{Accuracy results. The superscripts $^{+-=}$ next to the accuracies of PLML indicate the result of the McNemar's
statistical test with LMNN, BoostMetric, SML, CBLML, LMNN-MM, GMLM and SVM. They denote respectively a significant 
win, loss or no difference for PLML. The bold indicates the algorithm have a performance that is not significant different from the best algorithm. The number in the parenthesis indicates the score of the respective algorithm for the given dataset based 
on the pairwise comparisons of the McNemar's statistical test.}
\label{results}
\vskip 0.15in
 \scalebox{0.45}{
\begin{tabular}{l||c||ccc||ccc||c}
            &                                 & \multicolumn{3}{c||}{Single Metric Learning Baselines} & \multicolumn{3}{c||}{Local Metric Learning Baselines} & \\ \hline
Datasets    & PLML                            &LMNN                &BoostMetric &SML       &CBLML& LMNN-MM &GLML&SVM  \\ \hline \hline
Letter      &$\textbf{ 97.22}^{+++|+++|+}$(7.0) & 96.08(2.5)         & 96.49(4.5) & 96.71(5.5)& 95.82(2.5)& 95.02(1.0)& 93.86(0.0)& 96.64(5.0)\\
Pendigits   &$\textbf{ 98.34}^{+++|+++|+}$(7.0) & 97.43(2.0)         & 97.43(2.5) & 97.80(4.5)& 97.94(5.0)& 97.43(2.0)& 96.88(0.0)& 97.91(5.0)\\
Optdigits   &$\textbf{ 97.72}^{===|+++|=}$(5.0) &\textbf{ 97.55}(5.0)&\textbf{ 97.61}(5.0)&\textbf{ 97.22}(5.0)& 95.94(1.5)& 95.94(1.5)& 94.82(0.0)&\textbf{ 97.33}(5.0)\\
Isolet      &$\textbf{ 95.25}^{=+=|+++|=}$(5.5) &\textbf{ 95.51}(5.5)& 89.16(2.5)&\textbf{ 94.68}(5.5)& 89.03(2.5)& 84.61(0.5)& 84.03(0.5)&\textbf{ 95.19}(5.5)\\
USPS        &$\textbf{ 98.26}^{+++|+++|=}$(6.5) & 97.92(4.5)         & 97.65(2.5)& 97.94(4.0)& 96.22(0.5)& 97.90(4.0)& 96.05(0.5)&\textbf{ 98.19}(5.5)\\
MNIST       &$\textbf{ 97.30}^{=++|+++|=}$(6.0) &\textbf{ 97.30}(6.0)& 96.03(2.5)& 96.57(4.0)& 95.77(2.5)& 93.24(1.0)& 84.02(0.0)&\textbf{ 97.62}(6.0)\\
\hline \hline
 Total Score     &37&25.5&19.5&28.5&14.5&10&1&32.5 \\
\end{tabular}
}
\end{center}
\vskip -0.2in
\end{table}

\begin{figure}[t]

  \centering
  \subfigure[Letter]{\includegraphics[width=0.17\textwidth]{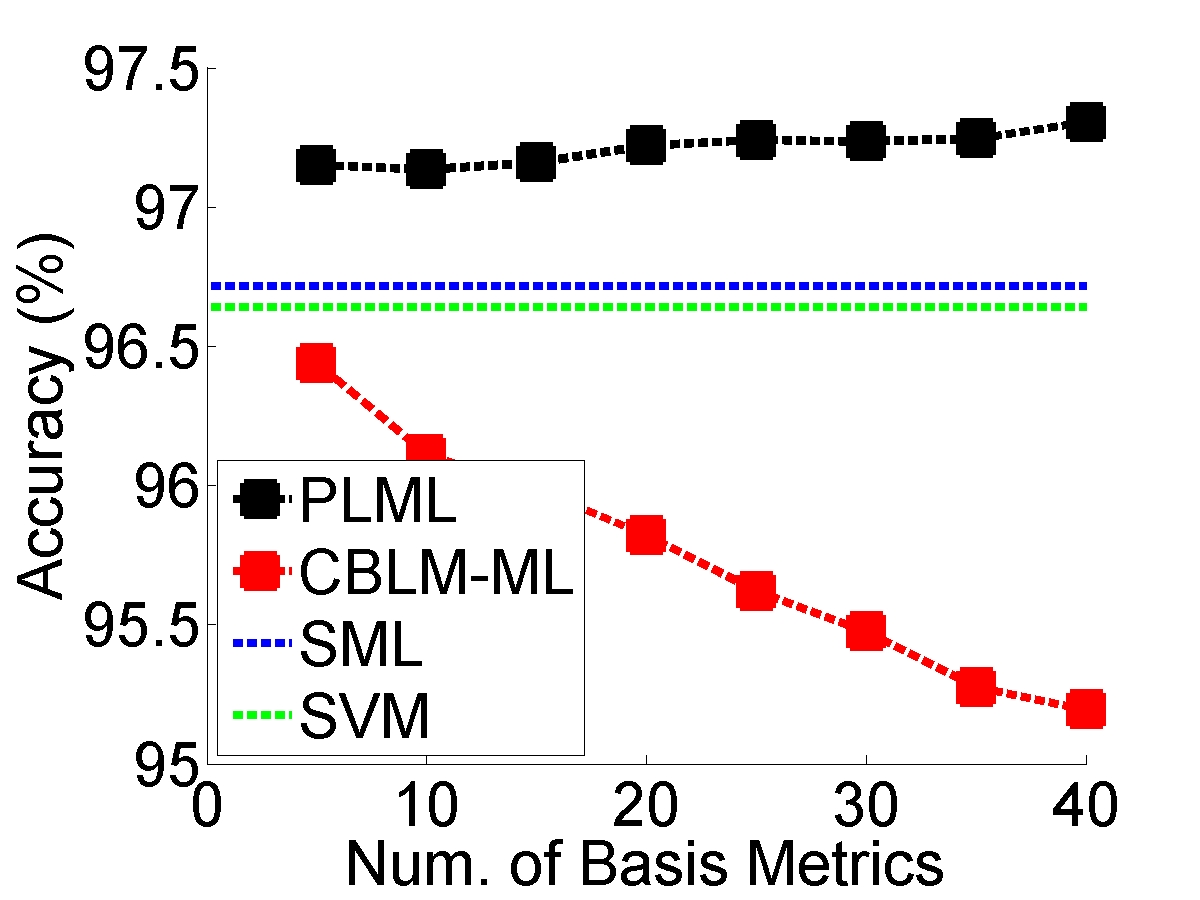}}
  \subfigure[Pendigits]{\includegraphics[width=0.17\textwidth]{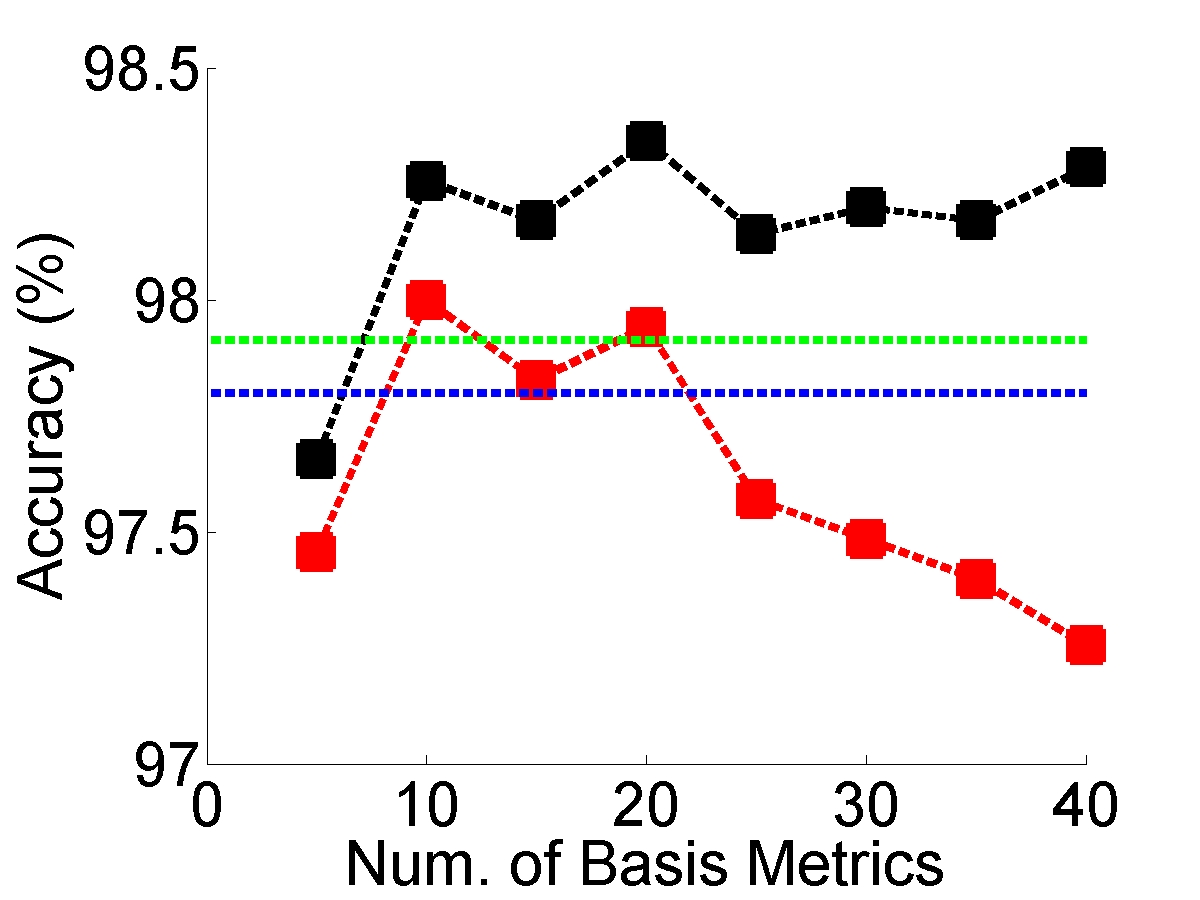}}
   \subfigure[Optdigits]{\includegraphics[width=0.17\textwidth]{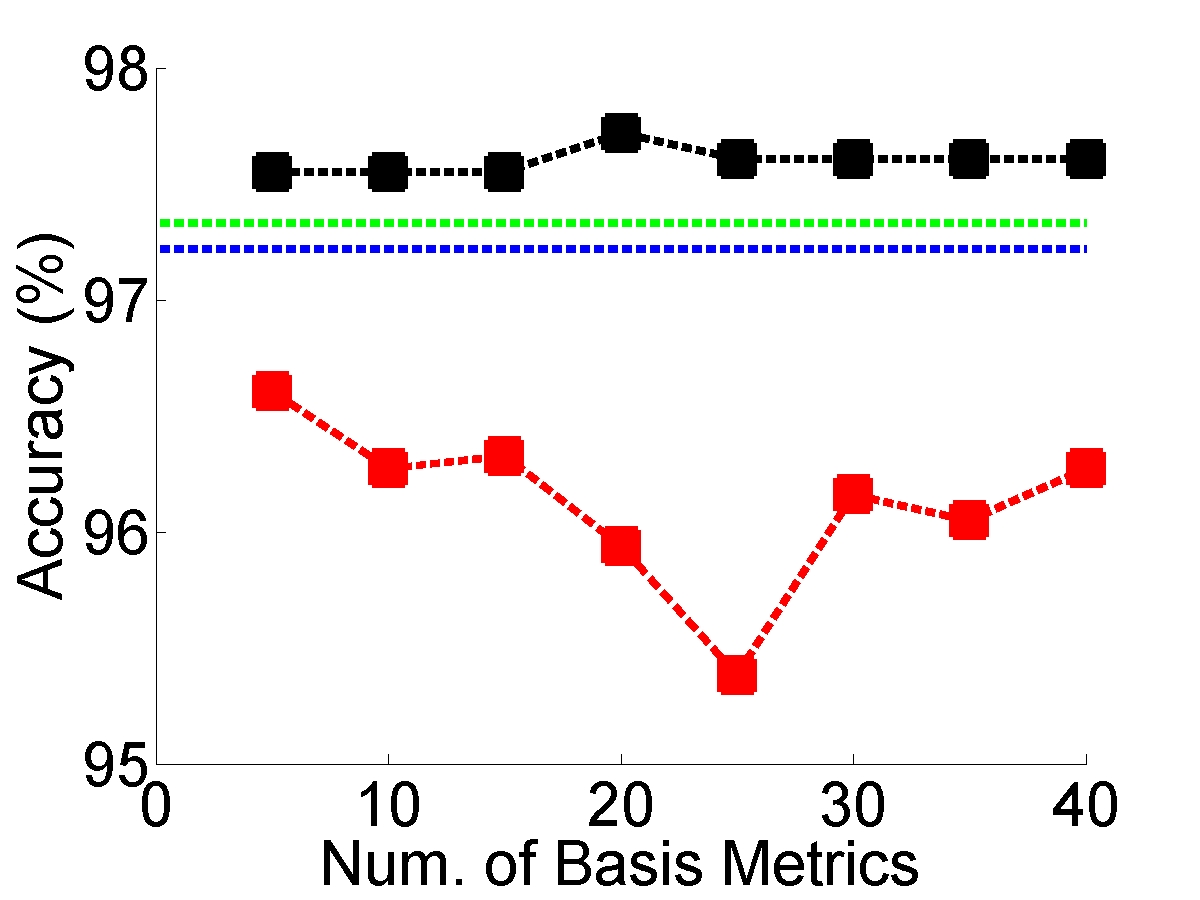}}
 
 \centering
  \subfigure[USPS]{\includegraphics[width=0.17\textwidth]{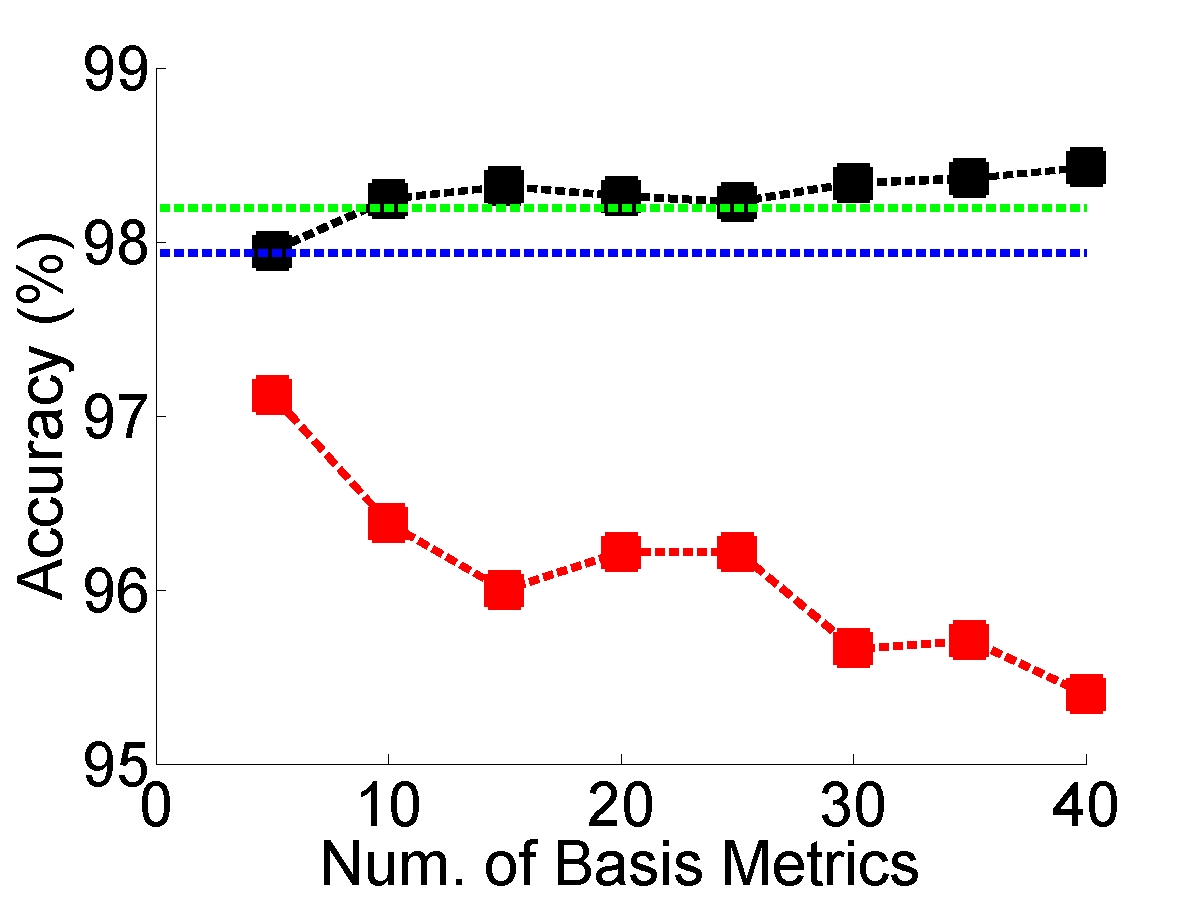}}
  \subfigure[Isolet]{\includegraphics[width=0.17\textwidth]{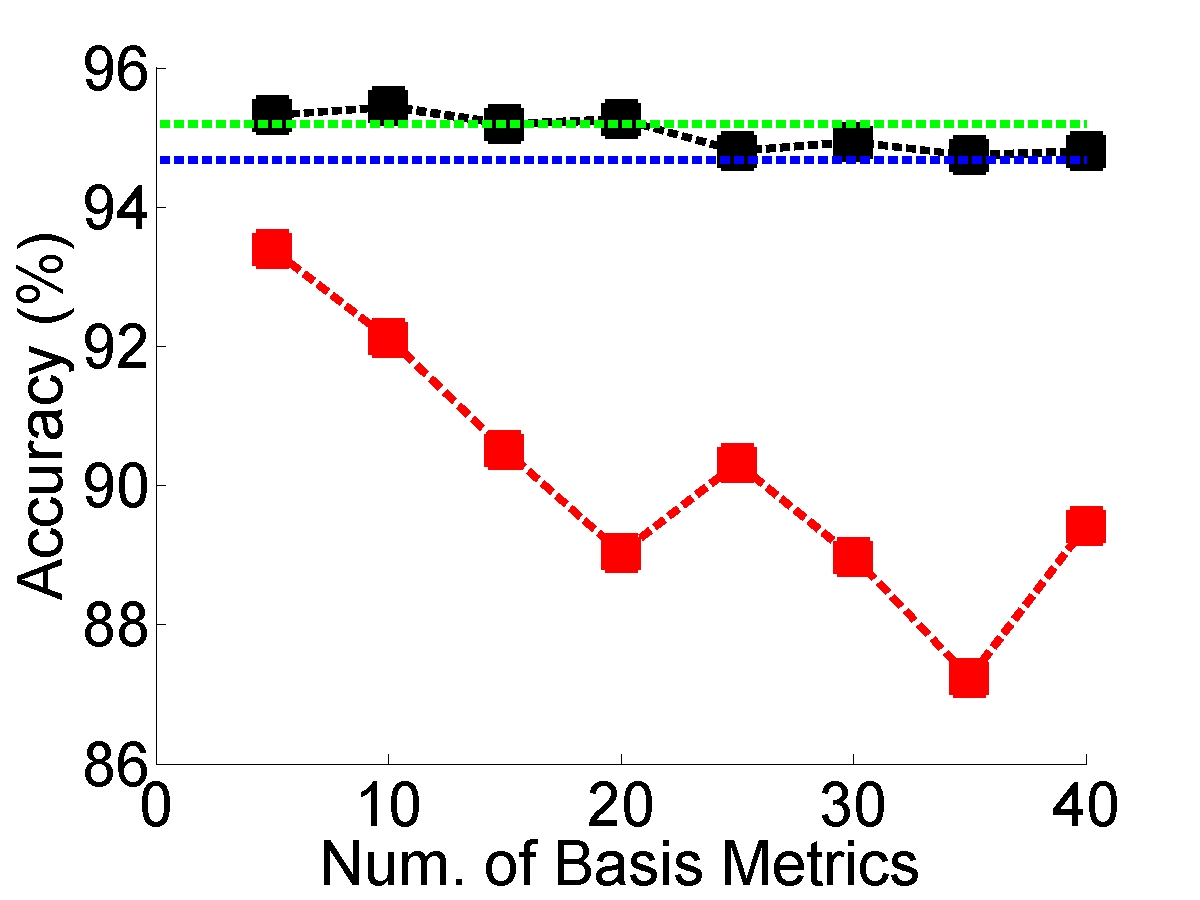}}
  \subfigure[MNIST]{\includegraphics[width=0.17\textwidth]{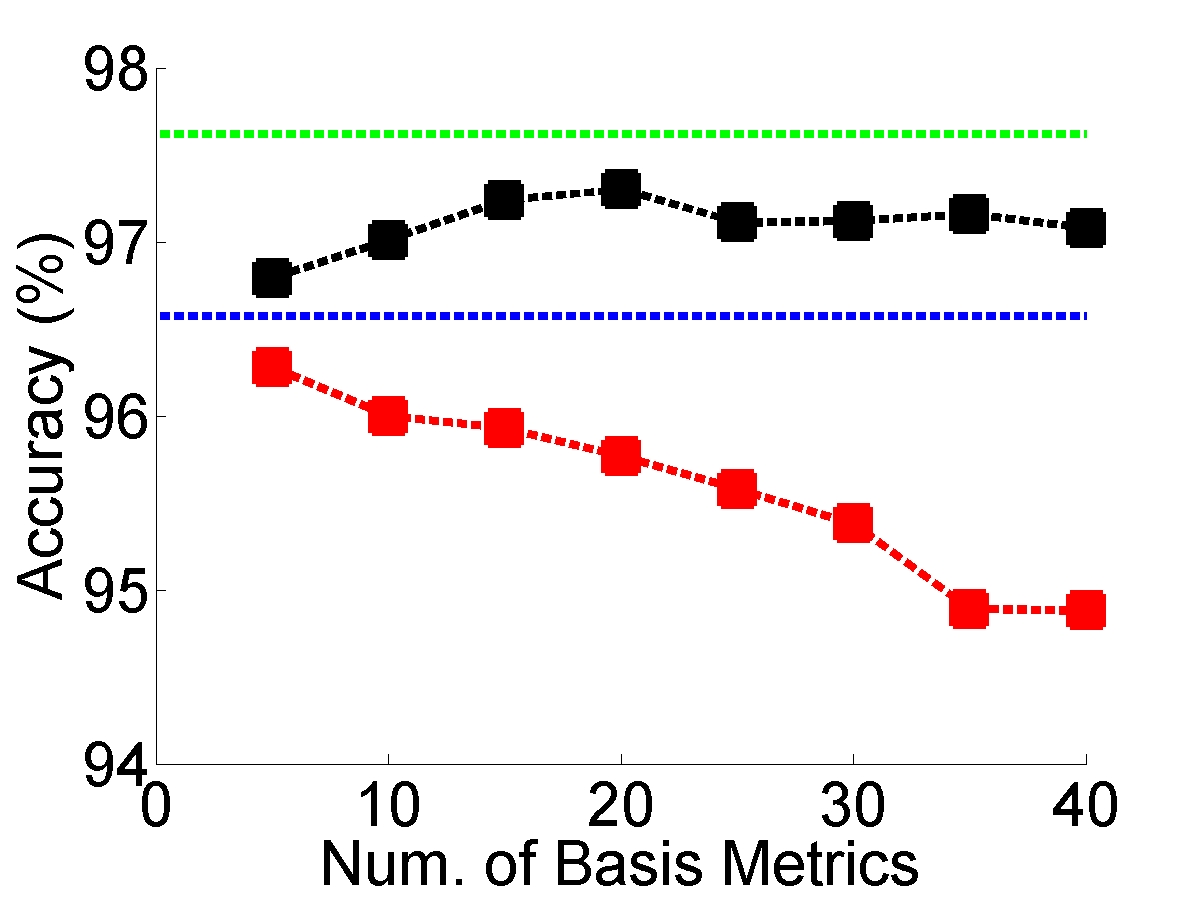}}
  \caption{Accuracy results of PLML and CBLML with varying number of basis metrics. }
\label{sensitivity}
\end{figure}
To estimate the classification accuracy for Pendigits, Optdigits, Isolet and MNIST we used the default train and test split, 
for the other datasets we used 10-fold cross-validation. The statistical significance of the differences were tested with McNemar's 
test with a p-value of 0.05. In order to get a better understanding of the relative performance of the different algorithms 
for a given dataset we used a simple ranking schema in which an algorithm A was assigned one point if it was found to have 
a statistically significantly better accuracy than another algorithm B, 0.5 points if the two algorithms did not have a 
significant difference, and zero points if A was found to be significantly worse than B.

\subsection{Results}
In Table~\ref{results} we report the experimental results. \PLML\ consistently outperforms the single global metric learning methods
LMNN, BoostMetric and SML, for all datasets except Isolet on which its accuracy is slightly lower than that of LMNN. Depending on 
the single global metric learning method with which we compare it, it is significantly better in three, four, and five datasets ( for LMNN, SML, and
BoostMetric respectively), out of the six and never singificantly worse. When we compare \PLML\ with CBLML and LMNN-MM, the two baseline methods which learn 
one local metric for each cluster and each class respectively with no smoothness constraints, we see that it is statistically 
significantly better in all the datasets. GLML fails to learn appropriate metrics on all datasets because its fundamental generative model 
assumption is often not valid. Finally, we see that \PLML\ is significantly better than SVM in two out of the six 
datasets and it is never significantly worse; remember here that with SVM we also do inner fold kernel selection to
automatically select the appropriate feature speace. Overall \PLML\ is the best performing methods scoring 37 points 
over the different datasets, followed by SVM with automatic kernel selection and SML which score 32.5 and 28.5 points 
respectively. The other metric learning methods perform rather poorly.
\note[Alexandros]{Why SML is so good, especially since it does not learn anything local, but just a single global metric?}

Examining more closely the performance of the baseline local metric learning methods CBLML and LMNN-MM we observe that they tend to 
overfit the learning problems. This can be seen by their considerably worse performance with respect to that of SML and LMNN 
which rely on a single global model. On the other hand \PLML\ even though it also learns local metrics it does not suffer from 
the overfitting problem due to the manifold regularization. The poor performance of LMNN-MM is not in agreement with the results 
reported in \cite{weinberger2009distance}. The main reason for the difference is the experimental setting. 
In~\cite{weinberger2009distance}, 30\% of the training instance of each dataset were used as a validation set to avoid overfitting.

To provide a better understanding of the behavior of the learned metrics, we applied \PLML\, LMNN-MM, CBLML and GLML, on an image dataset 
containing instances of four different handwritten digits, zero, one, two, and four, from the MNIST dataset. 
\note[Alexandros]{I think the comparison with  LMNN-MM here is unfair because it can only learn four metric matrices, 
a more appropriate might be with CBLML, or with LMNN-MM in which you also learn 20 metrics, as many as with \PLML. Or even
better simply include also CBML-ML?}
As in \cite{weinberger2009distance}, we
use the two main principal components to learn. Figure~\ref{visulization} shows the learned local metrics by plotting the axis 
of their corresponding ellipses(black line).  The direction of the longer axis is the more discriminative. Clearly 
\PLML\ fits the data much better than LMNN-MM and as expected its local metrics vary smoothly. In terms of the predictive performance, 
PLML has the best with 82.76\% accuracy. The CBLML, LMNN-MM and GLML have an almost identical performance with respective 
accuracies of 82.59\%, 82.56\% and 82.51\%.  
\note[Alexandros]{Doesn't this observation contradicts the overfitting statement done above? i.e. it seems that in 
any dataset LMNN-MM will rather underfit the data... Too many instances, few classes.}

Finally we investigated the sensitivity of \PLML\ and CBLML to the number of basis metrics, we experimented with 
$m \in \{5, 10, 15, 20, 25, 30, 35, 40\}$.
The results are given in Figure~\ref{sensitivity}. We see that the predictive performance of PLML 
often improves as we increase the number of the basis metrics. Its performance saturates 
when the number of basis metrics becomes sufficient to model the underlying training data. As expected different learning problems require different 
number of basis metrics. PLML does not overfit on any of the datasets. In contrast, the performance of CBLML gets worse when 
the number of basis metrics is large which provides further evidence that CBLML does indeed overfit the learning problems, demonstrating
clearly the utility of the manifold regularization.   

\section{Conclusions}
\label{sec:con}
Local metric learning provides a more flexible way to learn the distance function. However 
they are prone to overfitting since the number of parameters they learn can be very large. In this paper we presented \PLML, a local metric 
learning method which regularizes local metrics to vary smoothly over the data manifold. Using an approximation error bound of the metric 
matrix function, we parametrize the local metrics by a weighted linear combinations of local metrics of anchor points. Our method scales 
to learning problems with tens of thousands of instances and avoids the overfitting problems that plague the other local metric learning methods. 
The experimental results show that \PLML\ outperforms significantly the state of the art metric learning methods
and it has a performance which is significantly better or equivalent to that of SVM with automatic kernel selection. 



\subsubsection*{Acknowledgments}
This work was funded by the Swiss NSF (Grant 200021-137949). The support of EU projects DebugIT (FP7-217139) and e-LICO (FP7-231519), as well as that of COST Action BM072 ('Urine and Kidney Proteomics') is also gratefully acknowledged

\bibliography{LLMML}
\bibliographystyle{plain}
\end{document}